\documentclass{rsos}
\usepackage{booktabs}
\usepackage{xeCJK}
\Year{2026}
\hypersetup{hidelinks,pdfauthor={Peter Chen, Ni Zhang, Siyuan Zhang},pdftitle={Patching the Protocol: A Taxonomy of Pre-Registration Repairs in LLM-Proposer Experiments}}
\titlehead{Research}
\jname{rsos}
\Journal{R. Soc. Open Sci.\ }

\begin{document}

\title{Patching the Protocol: A Taxonomy of Pre-Registration Repairs in LLM-Proposer Experiments}

\author{Peter Chen$^{1,\dagger}$, Ni Zhang$^{1,\dagger}$ and Siyuan Zhang$^{1,*}$}

\address{$^{1}$Shenyang Institute of Technology, Shenyang, China}

\subject{computer science; statistics}


\corres{Siyuan Zhang\\
\email{lelandsy28@163.com}}

\others{$^{\dagger}$These authors contributed equally to this work.}

\begin{abstract}
Pre-registration is increasingly proposed as a trust mechanism for
machine-learning research involving LLM-generated content, but its
canonical form assumes a frozen protocol can be executed unchanged.
Software-intensive studies violate this assumption: the protocol is partly
implemented, and implementations carry defects. We report a three-case
study of protocol repairs (``patches'') from a program where an LLM
proposes executable rules that a programmatic verifier gates, across
production scheduling, incentive design, and task assignment. Each
protocol required exactly one defect-repair patch: three events in total.
None modified a numbered judgment criterion, its threshold, or the scoring
rule, and none altered already collected data; two changed pre-registered
design parameters before any data were collected, each with a recorded
rationale. We propose a two-axis taxonomy of patches (defect class:
environmental / feasibility / correctness, $\times$ timing: pre-data /
post-data), introduce a \emph{verifiable non-impact claim} for post-data
repairs, and distill a nine-item checklist; all nine items were exercised
by at least one patch event -- an observation of use, not a validation of
sufficiency. The credibility of verifier-bound, LLM-proposer studies
depends less on whether patches occur than on whether they are auditable.
\end{abstract}

\maketitle

\section{Introduction}
Discussions of trust in LLM-assisted research concentrate on a
dramatic failure mode: the model fabricates results~\cite{ji2023}, and the human passes them on. A quieter failure mode is more common in practice. Computational reproducibility has its own canon~\cite{peng2011,stodden2016,kapoor2023}, and replication campaigns in adjacent fields document how protocol drift undermines trust~\cite{osc2015,errington2021,camerer2018}; the broader case for pre-registering against such drift is long-standing~\cite{ioannidis2005,fanelli2012}.
Studies in which an LLM proposes \emph{code}~\cite{chen2021codex,gulwani2017,lu2024aiscientist} --- decision rules,
scoring functions, mechanisms --- are software-intensive, as the self-correction
literature that audits such programs attests~\cite{huang2024,madaan2023,shinn2023}: part of the
``protocol'' is a running program, and programs ship with defects
regardless of intent. A pre-registration that freezes hypotheses,
criteria, and budgets still gets executed by software that may crash
on boundary conditions, may encode mathematically infeasible
parameters, or may silently corrupt its own measurements.

The canonical pre-registration playbook --- OSF-style
freezing~\cite{nosek} and Registered Reports~\cite{chambers} ---
has no vocabulary for this. Its implicit model is: freeze, then execute. When execution
reveals that the protocol itself is defective, the researcher faces
an unforced choice --- run a broken study, or change the protocol and
appear to have moved the goalposts. Both outcomes damage credibility,
and the second one is indistinguishable from motivated tuning unless
the change is auditable.

Recent work extends pre-registration to LLM-based research --- to
experiments in which agents are the subjects \cite{prereggiagents2026}
and to LLM-generated data feeding downstream hypothesis tests
\cite{phacking2026} --- and audits LLM pipelines for accountability
layers that relay rather than check \cite{auditwithout2026} and for
silent API failure \cite{silentprobe2026}. None of these addresses the
case we study: a \emph{defect in the running protocol itself}, and how
to repair it without moving the goalposts.

The concern is not academic. A recent roadmap for recursive
self-improvement in research agents names reliable verification as
the bottleneck at every level of autonomy, and documents a live case
in which an agentic system cherry-picked favorable results during
automated experimentation \cite{theseus}. Verification
infrastructure, once deployed, becomes part of the protocol --- and
therefore part of what can break, and be repaired, mid-study.

This paper takes the position that \textbf{patches are normal and can
be made trustworthy}. We study three repairs that occurred in a
three-domain research program (production scheduling; incentive
mechanism design for mobile crowdsensing; capacity-constrained task
assignment). Each domain ran under a pre-registered
protocol~\cite{compA,compB,compC} with
append-only changelogs, frozen judgment criteria, negative-result
clauses, leakage guards, and mandatory equal-budget random controls.
Each protocol required exactly one defect-repair patch; across all
three, no patch modified a numbered criterion, its threshold, the
scoring rule, or any collected data point (two patches did change
pre-registered design parameters before data collection; \S3 gives
the component-by-component verdicts).

\textbf{Contributions.}
\begin{enumerate}
  \item \textbf{A three-case taxonomy of protocol repairs},
  classified along two axes: defect class (environmental /
  feasibility / correctness) and timing (pre-data / post-data).
  \item \textbf{The concept of a verifiable non-impact claim}: for
  post-data repairs, the changelog must not merely assert that
  already collected data are unaffected, but must contain a check
  that a reader can re-run.
  \item \textbf{A nine-item checklist} for disciplined protocol
  repair in software-intensive ML studies, distilled from the cases
  and shipped as a reusable pre-registration template; all nine items
  are exercised by at least one case (nine of nine;
  Table~\ref{tab:matrix}).
\end{enumerate}

The checklist is not externally validated (\S7); our claim is
narrower --- that these practices made three specific repairs
auditable, and that the audit trail is inspectable in the released
archives.

\section{Background: the protocol layer}

\subsection{Setting}
The research program uses a verifier-bound design: an LLM proposes
executable rules; a programmatic verifier checks candidate rules
against frozen golden properties; a held-out evaluation scores
surviving rules. Three domains were studied sequentially:

\begin{itemize}
  \item \textbf{Domain A (production scheduling)}: the LLM proposes
  job-to-slot assignment rules; the verifier checks feasibility and
  overdue constraints; the outcome measure is new-overdue cost on
  held-out instances \cite{compA}.
  \item \textbf{Domain B (incentive design)}: the LLM proposes
  payment functions \texttt{pay(report\_c, report\_q, signal, round,
  budget\_left)}; the verifier enumerates worker types and misreport
  grids to check individual rationality (IR) and incentive
  compatibility (IC); the primary outcome is the fraction of accepted
  mechanisms \cite{compB}.
  \item \textbf{Domain C (task assignment)}: the LLM proposes
  scoring functions over (distance, capability, window, load, \dots);
  the verifier caps coverage degradation and inequality (Gini); the
  outcome is a composite $J$ \cite{compC}.
\end{itemize}

\subsection{Protocol-layer components}
Every domain ran under the same five protocol components, frozen
before data collection:

\begin{enumerate}
  \item \textbf{Frozen criteria.} Each pre-registration lists
  numbered criteria with thresholds and pre-committed result
  branches; adding, modifying, or deleting criteria after data
  collection is prohibited; exploratory analyses must be labeled as
  such.
  \item \textbf{Append-only changelog.} Any protocol change is a
  dated patch appended after existing entries, which are preserved
  verbatim. Each patch declares its effect on already collected data.
  \item \textbf{Negative-result clauses.} Failure branches and
  ``gray-band'' outcomes are pre-committed; gray bands may not be
  narrated as wins.
  \item \textbf{Leakage discipline.} Held-out data are constructed by
  physically separate code; each promoted rule (baselines included)
  is evaluated on the holdout exactly once, enforced by a counter; a
  runtime probe test asserts that no evaluation module can read
  ground-truth keys even if they are injected into the context.
  \item \textbf{Equal-budget controls.} Every LLM condition is paired
  with a random-proposer control spending the same evaluation budget
  (evals are the budget currency), with control seeds independent of
  instance seeds.
\end{enumerate}

Component 2 is the subject of this paper: it is what turns a repair
from a credibility hazard into an auditable event.

\section{Three cases}
\textbf{Counting rules.} All counts in this paper are patch-event
counts under the following conventions, fixed before writing. A
\emph{substantive repair} is one changelog entry that records a
defect-repair patch, however many individual changes that entry
bundles; entries that freeze a draft into v1.0, extend the
pre-registration with a new study stage, or adjust reporting
wording are not defect repairs and are not counted. A patch
\emph{touches} a judgment criterion, threshold, or scoring rule if
it changes the numbered criterion text, the numeric threshold or
result branch attached to it, or the rule that computes the outcome
measure; changes to the surrounding machinery that make a frozen
criterion evaluate as specified (or to pre-registered design
parameters such as budgets and task counts) are recorded separately
and do not count as touching. The full changelog histories are: Domain~A, v1.0 $\to$ v2 (Stage-2 pre-registration added) $\to$ v3
(Stage-4 pre-registration added) $\to$ v3.1 (the repair of
Case~1; one repair event); Domain~B, v0.1 $\to$ v1.0 (freeze) $\to$
v1.1 (the repair of Case~2; one repair event bundling four
sub-changes) $\to$ v1.2 (reporting-wording adjustment; not a
repair); Domain~C, v0.1 $\to$ v1.0 (freeze) $\to$ v1.1 (the repair
of Case~3; one repair event bundling three sub-changes). Each case's
repair section walks its own sub-changes so that the headline counts
can be re-derived component by component.

Each case follows the same template: archive $\to$ discovery context
$\to$ defect $\to$ repair $\to$ effect on collected data $\to$
classification. Changelog text is quoted verbatim (translated where
necessary). Table~\ref{tab:cases} summarizes the scale of each domain
and the status of its data at patch time; all numbers are taken from
the frozen pre-registrations and append-only changelogs.

\begin{table}[t]
\centering\small
\setlength{\tabcolsep}{4pt}
\begin{tabular}{@{}p{0.10\linewidth}p{0.16\linewidth}p{0.21\linewidth}p{0.30\linewidth}p{0.11\linewidth}@{}}
\toprule
Domain & Discovery context & Scale at patch time & Patch (key parameters) & Data at patch \\
\midrule
A & run-time crash at pair 037 & batch study; 37 pairs written &
redraw rule extended train $\to$ holdout (seed$+1$) & 37 pairs
retained, re-validated \\
B & 1-pair smoke test & $N=20$ workers; budget $B=50$ & $B$: $50 \to
100$; participation constraint; baseline exemption & none
(pre-data) \\
C & smoke debugging & capacity $30 \times 3 = 90$; $M \in
\{50,80,100\}$ & deep-copy/idempotent \texttt{run\_match}; $M$:
$50 \to 60$; deadlines $10$--$30 \to 8$--$15$ min & none (pre-data) \\
\bottomrule
\end{tabular}
\caption{Per-domain scale, patch, and data status at patch time.
Pair indices are zero-based (pairs 0--36 complete, crash at 037).
Symbols ($B$, $M$, capacity) are defined in the case narratives and
\S2.1. A smoke test is a minimal pre-experiment trial run whose
outputs are not data.}
\label{tab:cases}
\end{table}

\subsection{Case 1 (Domain A): environmental defect --- crash at a
boundary}
\textbf{Archive.} Domain A pre-registration, changelog v3.1
(2026-09-11).

\textbf{Discovery context.} Run time. The batch study crashed at pair
037; 37 pairs had already been written to disk.

\textbf{Defect.} The instance-generation procedure \texttt{draw\_pair}
had a degenerate-case redraw rule that covered only \emph{training}
instances. When the holdout baseline came up infeasible (no feasible
schedule within 40 days), generation crashed instead of redrawing.

\textbf{Repair} (verbatim, Chinese): ``draw\_pair
的退化重抽规则从仅 train 扩展到 holdout（holdout baseline
INFEASIBLE 或排产 $>$40 天时 seed+1 重抽）。此为崩溃修复而非口径
变更，与 v3 预注册『退化重抽 seed 递增』精神一致'' --- the redraw
rule of \texttt{draw\_pair} was extended from train-only to holdout
as well: when a holdout baseline is INFEASIBLE or schedules beyond
40 days, redraw with an incremented seed. This is a crash fix, not a
criteria change, and is consistent with the pre-registered
``degenerate-case seed-increment redraw'' policy of pre-registration
v3 (the changelog entry is v3.1; v3 there denotes the
pre-registration version). The
extension is not merely consistent with that policy --- the frozen
text already required it: ``if any instance's baseline evaluation is
INFEASIBLE or schedules beyond 40 days, redraw with an incremented
seed and record the redraw count''; the rule is stated over
\emph{any} instance, train or holdout. The defect was that the
implementation covered only training instances; the patch aligned
the code with the policy it was supposed to implement.

\textbf{Effect on collected data.} The changelog carries a
\emph{verifiable non-impact claim}: ``前 37 对（pair 0--36）在旧逻辑
下产出，其 holdout baseline 均已验证可行，数据保留有效'' --- each of
the 37 written pairs was individually checked and its holdout baseline
confirmed feasible, so the old-logic data are retained. The check is
mechanical: a reader can re-run feasibility validation on pairs
0--36. The claim's scope is exactly what the check computes: the
non-impact claim asserts that the holdout baselines of pairs 0--36
are feasible under the frozen instance specification, and the check
recomputes precisely that quantity; the check is a deterministic
recomputation over frozen instance snapshots --- the same inputs the
verifier consumed at collection time --- so a reader re-running it
obtains the changelog's verdict directly, and the snapshots and the
check's code ship in the Domain~A archive.
Two boundaries are stated explicitly. First, the claim does not
cover the prospective composition of pairs 037 onward: because the
redraw policy conditions instances on baseline feasibility, the
holdout population from the repair point onward is
feasibility-conditioned --- which is what pre-registration v3
already intended, but a reader comparing raw draws to
post-patch draws should know the conditioning applies. Second, the
checker is itself program code inside the protocol and shares its
failure modes; the program therefore requires non-impact checks to
be minimal, independently testable code whose own defects are
changelog-managed like any other patch (checklist items 2 and 9).

\textbf{Classification.} Environmental: the defect is a crash in the
data-generation pipeline at a boundary condition. Symptoms are
visible (the run stops), so the risk of unnoticed damage is low. The
repair does not alter what is measured or accepted.

\subsection{Case 2 (Domain B): feasibility defect --- infeasible
parameters, intercepted at design time}
\textbf{Archive.} Domain B pre-registration, changelog v1.0 $\to$
v1.1 (2026-09-12), after a one-pair smoke test.

\textbf{Discovery context.} Design time. A single smoke pair failed
to produce any verifier-acceptable mechanism, prompting a by-hand
audit of the parameter regime.

\textbf{Defect.} The budget made IR and the budget cap mathematically
incompatible. In the simulated market, worker $i$'s utility from a
matched task is $u_i = \mathrm{pay}_i - c_i d_j$: the payment minus
the effort cost of completing the assigned task. The verifier checks
IR by requiring honest utility $\ge 0$ for all $N=20$ enumerated
worker types, and IC by enumerating a misreport grid per type
($\mathrm{report\_c}$ at 11 points, $\mathrm{report\_q}$ at 5
points, $\epsilon = 0.05$). A payment that is IC-safe in this design
cannot profitably depend on the reports: any report dependence a
worker could exploit is caught by the grid, so the binding payment
structure is a constant. Individual rationality for the worst worker
type (cost $c=2.0$, effort $d=1.5$) then requires a constant payment
of at least $c \cdot d = 3.0$ per worker. With $N=20$ workers the
minimum total outlay is $20 \times 3.0 = 60 > B = 50$. No mechanism
can satisfy IR and the budget simultaneously: the verifier's
acceptance rate is structurally zero for reasons that have nothing
to do with any proposer, human or machine.

\textbf{Repair.} Budget $B$: $50 \to 100$, restoring a non-empty
feasible region (the changelog's recorded rationale; $B=100$ leaves
the binding floor of $60$ non-trivial headroom). The same patch,
from the same smoke test, fixed three adjacent design defects:
(i)~a participation constraint was added to the simulation, so that
IR violations have real economic consequences (without it, ``compliant mechanisms pay more'' loses to violating mechanisms by construction, contradicting the
pre-registered criterion); (ii)~baselines are exempt from the
verifier gate --- they are reference lines, and their verifier status
is reported as evidence, not used for exclusion; (iii)~the prompt was
hardened to eliminate irrelevant parse failures.

\textbf{Component verdicts.} Under the counting rules above, none of
the four sub-changes touched a criterion. (i) changes the market
semantics of the simulation: the frozen primary criterion (LLM-best
holdout welfare vs.\ the three baselines) is unchanged in text and
threshold, but without participation the simulated welfare ordering
contradicts the economics that criterion assumes --- the patch makes
the frozen criterion evaluate as specified rather than redefining
it. (ii) changes gate scope: which evaluated objects must pass the
verifier before entering the comparison; the comparison structure of
the criterion is unchanged, and all three baselines still enter the
holdout evaluation. (iii) is a pipeline-behavior fix at the parse
boundary. The budget change is a pre-registered design parameter
chosen pre-data; its selection procedure is the changelog's recorded
rationale, and we flag in \S7 that repair-parameter selection
remains a discretion this checklist does not yet cover.

\textbf{Effect on collected data.} None: the experiment had not
started; the smoke pair was discarded by rule (checklist item 8).

\textbf{Classification.} Primary class: feasibility --- the defect
lives in the \emph{specification}, not the code: an infeasible
parameter regime that any smoke test could expose. This is the
cheapest class to catch and the most damaging to miss: an unnoticed
infeasible regime produces a structurally zero acceptance rate that
could be misread as a scientific finding (``LLMs cannot design
mechanisms'') rather than an artifact. Secondary class: correctness
--- sub-change (i) alters measurement semantics (welfare accounting)
so that the frozen criterion is evaluated as specified, which is a
correctness-class change in our own taxonomy; the primary class is
the one whose damage profile would dominate had the defect been
missed. Sub-change (iii) is environmental in character (a boundary
behavior of the generation pipeline).

\subsection{Case 3 (Domain C): correctness defect --- silent
measurement corruption}
\textbf{Archive.} Domain C pre-registration, changelog v1.0 $\to$
v1.1 (2026-09-12), after smoke debugging; the fix is documented in
the simulation module's docstring.

\textbf{Discovery context.} Smoke debugging, pre-data.

\textbf{Defects (two, one code-level and one design-level).}
\emph{Code level.} \texttt{run\_match} mutated the evaluation batch in
place. When several mechanisms were scored against the same batch,
each mechanism's run contaminated the state seen by the next; the
observable symptom was that all mechanisms received identical
composite scores ($J$). Note what makes this the most dangerous
class: the code \emph{succeeds} --- no crash, no exception,
plausible-looking numbers. Had the study run, it would have produced
a full dataset in which no mechanism can be distinguished from any
other, and nothing about the output would flag it.

\emph{Design level.} The task load had no region of genuine tension.
With $M=50$ tasks the coverage term is trivially saturated; with
$M=80$ or $100$, tasks exceed total capacity ($30 \times 3 = 90$),
every worker ends up fully loaded, and the Gini fairness term is
identically zero. All three tested values lie in structurally
degenerate parameter regions.

\textbf{Repair.} \texttt{run\_match} now deep-copies internally and
is idempotent (documented and unit-testable); task count $M$: $50 \to
60$, and deadlines $10$--$30 \to 8$--$15$ minutes, placing the regime
in the intermediate band where coverage is non-trivial and time
windows bind. The changelog records the selection procedure as an
explicit three-region analysis over the pre-registered candidates:
$M=50$ saturates the coverage term (trivially easy); $M \in
\{80, 100\}$ exceeds total capacity ($30 \times 3 = 90$), so every
worker ends up fully loaded and the Gini term is identically zero;
the intermediate value $M=60$ leaves a finite but non-empty set of
time-feasible assignments, which is the only region in which the
coverage--fairness tradeoff has genuine tension. The deadline band
$8$--$15$ minutes was chosen in the same analysis so that the
time-feasibility constraint actually binds at $M=60$. As with the
Domain~B budget, we flag in \S7 that this parameter selection was
author judgment, documented post hoc in the changelog rather than
pre-committed by rule.

\textbf{Effect on collected data.} None: the experiment had not
started; the smoke pair was discarded.

\textbf{Classification.} Primary class: correctness (code level) ---
a silent defect in measurement semantics. It can only be caught by
design review, idempotency discipline, or the luck of a smoke-test
symptom --- never by monitoring the run. Secondary class:
feasibility (design level) --- the degenerate parameter regions.
Both classes are labeled in the changelog entry (\S4.1).

\section{Taxonomy}

\subsection{Axis 1: defect class}

\begin{table}[t]
\centering\small
\setlength{\tabcolsep}{4pt}
\begin{tabular}{@{}p{0.18\linewidth}p{0.19\linewidth}p{0.16\linewidth}p{0.19\linewidth}p{0.17\linewidth}@{}}
\toprule
Class & Locus & Symptom & Detection cost & Damage if missed \\
\midrule
Environmental & pipeline boundary conditions & crash (loud) & low,
stops the run & lost compute, recoverable \\
Feasibility & specification / parameters & structurally degenerate
outcomes & low, one smoke cycle & misread artifact as finding \\
Correctness & measurement semantics & none (silent) & high, needs
audit discipline & invalid dataset with plausible numbers \\
\bottomrule
\end{tabular}
\caption{The defect-class axis. Detection-cost and damage entries for
feasibility and correctness rest on one observed instance each; treat
them as working hypotheses, not estimates.}
\label{tab:axis1}
\end{table}

The classes are ranked by \emph{audibility}
(Table~\ref{tab:axis1}). Environmental defects
announce themselves; feasibility defects announce themselves through
degenerate outcome distributions if one looks; correctness defects
say nothing. Protocol design should therefore spend its review budget
in inverse proportion to audibility: unit-testable idempotency and
state-isolation invariants (correctness), smoke-test regime audits
(feasibility), and simple redraw robustness (environmental).

The classes describe the \emph{primary} defect of a patch, but a
single smoke cycle can surface defects in more than one class; Case 3
pairs a correctness defect (code level) with a feasibility defect
(design level), and Case 2 adds a correctness-class sub-change to a
feasibility-class patch. Such patches carry a primary and a secondary
class in the changelog, and the primary class is the one whose damage
profile would dominate had the defect been missed.

\subsection{Axis 2: timing}
\begin{itemize}
  \item \textbf{Pre-data repairs} (Cases 2, 3) enjoy maximum freedom:
  the cost of a patch is a discarded smoke pair. Nothing else is
  owed.
  \item \textbf{Post-data repairs} (Case 1) owe the archive a
  \emph{verifiable non-impact claim}: not ``these data are fine'' but
  ``here is the check, here are the 37 pairs, re-run it yourself.''
  The claim in Case 1 is mechanical (feasibility validation of pairs
  0--36) and cheap to re-verify.
\end{itemize}

\subsection{The red line}
\textbf{Case inclusion.} The three cases are all of the program's
\emph{domain} pre-registrations that existed when this paper was
written: each produces its own domain data under a frozen protocol.
The program also runs cross-domain \emph{analysis} artifacts, which
consume domain results rather than produce domain data; those follow
the same changelog discipline but are outside this paper's case set
by construction, not by curation.

\textbf{The red line.} The taxonomy deliberately excludes one
category: \textbf{post-data patches that alter criteria, thresholds,
or scoring}. In the domain protocols such changes are prohibited
outright. One cross-domain analysis artifact did require a protocol
revision: its shared bucketing rule (quartile buckets over positive
train--holdout margins) was ill-defined for a zero-gain population,
so the analysis protocol was revised to v1.1, restricting the shared
buckets to pairs with positive margin; the revision was declared
\emph{before} the analysis was re-run and recorded in the artifact's
changelog \cite{comp4}. We report this episode not as a counterexample
but as the boundary case it is: the revision touched an analysis
protocol, not a domain criterion, and it is the kind of change the
red line would prohibit inside a domain protocol mid-study. The
taxonomy's purpose is to make this boundary legible: patches that
repair the vehicle are normal; patches that steer toward a desired
result are not, and the two are separated by timing, by what the
patch touches, and by which ledger records it --- all visible in the
archives.

\textbf{Why the red line is asymmetrical.} The three defect classes
differ sharply in what a post-data repair can honestly promise. An
environmental post-data repair can carry a mechanical non-impact
claim (Case 1: feasibility re-validation of the 37 written pairs) ---
this is the one observed instance. The other two arms of the
asymmetry are analytic conjectures, not observations, and should be
read as such. A feasibility post-data repair might plausibly carry a
non-impact claim, since specification defects tend to produce visibly
degenerate outcome distributions, but the study contains no
post-data feasibility instance to confirm it. A correctness
post-data repair generally cannot carry one: a defect in measurement
semantics silently invalidates the data it produced, and no per-pair
check can certify measurements that were computed wrongly from the
start. This asymmetry is the reason the red line is drawn where it
is. The classes whose post-data repairs remain verifiable are exactly
the classes that do not touch scoring; any post-data change that does
touch scoring is therefore presumptively unverifiable, which is why
the domain protocols prohibit it outright rather than merely
documenting it.

\section{The checklist}
Shipped as a fill-in template; reproduced here in full.

\begin{enumerate}
  \item \textbf{Freeze criteria before data.} Numbered criteria,
  thresholds, result branches. No post-hoc additions, modifications,
  or deletions; label exploratory analyses as exploratory.
  \item \textbf{Append-only changelog.} Prior entries preserved
  verbatim; every patch declares its effect on collected data
  (``none'' + reason, or a re-runnable check).
  \item \textbf{Negative-result clauses.} Pre-commit failure branches
  and gray-band narratives; gray bands may not be narrated as wins.
  \item \textbf{Leakage discipline.} Holdout built by separate code;
  holdout evaluated exactly once per rule (counter-enforced); runtime
  probe tests assert no module can read ground-truth keys.
  \item \textbf{Equal-budget controls.} Report a same-evals
  random-proposer control for every LLM condition; control seeds
  independent of instance seeds.
  \item \textbf{Log every rejection verbatim} (sandbox and parse
  failures).
  \item \textbf{Three-state verifiers.} If a property cannot be
  evaluated, the verdict is \emph{uncertain}, never a silent pass.
  \item \textbf{Smoke pairs are not data.} Smoke-test outputs are
  discarded by rule, which is what makes design-time patches cheap.
  \item \textbf{Classify every patch} --- environmental / feasibility
  / correctness $\times$ pre-data / post-data --- in the changelog
  entry itself.
\end{enumerate}

Items 1--3 follow standard pre-registration practice; items 4--8 are
specific to software-intensive studies; item 9 is this paper's
addition.

\textbf{The 9/9 claim, made checkable.} A checklist item is
\emph{exercised} in a case if that case's protocol, changelog, or
archive shows the practice instantiated and load-bearing for the
study or the repair. Item~$k$ is counted if at least one case
exercises it. Table~\ref{tab:matrix} gives the per-item mapping; the
abstract's nine-of-nine count can be checked against it without
leaving the paper. One honesty note: this checklist was distilled
\emph{from} the cases, so citing an item number inside a case
narrative (as Case~2 does for item 8) is retrospective codification,
not a claim that the numbered rule governed the episode at the time;
the underlying practice did, as the cited changelog text shows.

\begin{table}[t]
\centering\small
\setlength{\tabcolsep}{4pt}
\begin{tabular}{@{}p{0.30\linewidth}p{0.20\linewidth}p{0.22\linewidth}p{0.20\linewidth}@{}}
\toprule
Checklist item & Domain A & Domain B & Domain C \\
\midrule
1 Freeze criteria & pre-reg v1--v3 freeze numbered criteria & v1.0
freeze & v1.0 freeze \\
2 Append-only changelog & v1$\to$v2 $\to$ v3 $\to$ v3.1 &
4 entries, v0.1 to v1.2 & 3 entries, v0.1 to v1.1 \\
3 Negative-result clauses & failure branches pre-committed & failure
branches pre-committed & failure branches pre-committed \\
4 Leakage discipline & holdout evaluated once, counter-asserted &
\texttt{HoldoutGate} & \texttt{HoldoutGate} \\
5 Equal-budget controls & 20 random proposals per pair & 20 random
proposals per pair & 20 random proposals per pair \\
6 Log rejections verbatim & parse-fail rate archived; bad proposals
kept & violation ledger fed back & violation ledger fed back \\
7 Three-state verifiers & sandbox exception = ERROR candidate, not
crash & baseline verifier status recorded & baseline verifier status
recorded \\
8 Smoke pairs not data & n/a (no smoke pair; run-time case) & 1 pair
discarded by rule & 1 pair discarded by rule \\
9 Classify every patch & v3.1: ``crash fix, not a criteria change''
& v1.1: infeasible parameter regime & v1.1: silent corruption +
degenerate regime \\
\bottomrule
\end{tabular}
\caption{Checklist items $\times$ cases. Every item is exercised in
at least one case (nine of nine). Cell contents are the archive
facts; item 8 is not applicable in Domain A, which had no smoke
pair.}
\label{tab:matrix}
\end{table}

\section{Discussion}
\textbf{Relation to existing practice.} The discussion continues,
rather than repeats, the positioning sketched in the introduction.
OSF-style pre-registration and Registered Reports~\cite{nosek,chambers} freeze \emph{hypotheses and analysis plans};
neither anticipates that the execution vehicle is code with its own
defect surface. Software engineering's response to the same problem is regression-style gatekeeping~\cite{ammann2017,barr2015}: changes are allowed, but each change carries
a rationale, a scope, and a test. The changelog transplants that
discipline into the pre-registration: a changelog entry is a commit
message, the ``effect on collected data'' field is the test, and the
append-only rule is immutability.

Pre-registration practice already has a vocabulary for protocol
change: OSF-style deviation documentation and the Registered Reports
Stage-2 rules distinguish minor deviations from major amendments, and
pre-specified analyses from exploratory ones \cite{nosek,chambers}. Our two axes are complementary rather than
competing. The timing axis refines the amendment point: pre-data
patches sit closer to design-time specification changes than to
deviations from a running study. The defect-class axis adds a
dimension those frameworks lack --- the locus of the change in the
execution vehicle --- and that locus is what determines what kind of
non-impact evidence is possible at all.

\textbf{Patches did not lower the bar.} Across the three cases, no
numbered criterion, threshold, or scoring rule changed; the one
post-data repair extended an existing, pre-registered policy
(redraw-on-degenerate) to a split it had not covered, and the
pre-registration text already stated that policy over \emph{any}
instance (\S3, Case~1). Two pre-data patches did change
pre-registered design parameters (the Domain~B budget and gate
scope; the Domain~C task count and deadline band), each with its
rationale recorded in the changelog and walked component by
component above. The changelogs are deposited as versioned Zenodo
records, so the patch history carries third-party deposit
timestamps; a skeptical reader can check each claim in this paper
against those records --- within the limits of a self-audited
program (\S7).

\textbf{What the taxonomy buys.} We found naming the classes
useful in our own review discussions, though this is self-report
rather than evidence. ``The protocol was patched'' is an accusation; ``a
feasibility-class, pre-data patch, with the smoke pair discarded'' is
a status report. We found the \emph{correctness} class the most
sobering: Case 3's defect would have produced a complete,
well-formed, uninformative dataset, and only the smoke-test symptom
and the idempotency review stood between it and the archive.

\section{Limitations}
Three cases from one program, classified by the program's own
authors: this is a self-audit, not an independent one, and $n=3$
supports the taxonomy's \emph{plausibility}, not its completeness ---
other defect classes may exist. The checklist has not been adopted or
stress-tested by an external team. The underlying studies are
simulations, not human-subject experiments; the protocol layer
transfers, but the stakes profile does not. Finally, verifiable
non-impact claims are only as good as their checks; a program that
always writes ``no impact'' without a re-runnable check has adopted
the vocabulary of this paper without its substance. Finally, the
selection of new parameter values in pre-data repairs (the
Domain~B budget, the Domain~C task count and deadline band) was
author judgment recorded in the changelog, not a pre-committed
selection rule; a checklist that disciplines repair \emph{reporting}
does not yet discipline repair \emph{parameterization}, and we
regard that as an open design problem.

\section{Conclusion}
Software-intensive research cannot promise defect-free protocols; it
can promise auditable repairs. A two-axis taxonomy (environmental /
feasibility / correctness $\times$ pre-data / post-data), a changelog
that carries verifiable non-impact claims, and a nine-item checklist
turn ``we changed the protocol'' from a credibility hazard into a
status report --- for verifier-bound, LLM-proposer studies today,
and for closed-loop automated research tomorrow, where the same
trust question arrives with higher stakes. The archives of the three studies --- pre-registrations
with append-only changelogs, result ledgers, and analysis scripts ---
are released so that every claim in this paper can be checked against
the patch record it describes.

\dataccess{The pre-registration documents (with append-only changelogs), run ledgers including rejected proposals, analysis scripts and failure records are archived on Zenodo (per-domain records: 10.5281/zenodo.23433811, 10.5281/zenodo.23433817, 10.5281/zenodo.23433820; cross-domain analysis record: 10.5281/zenodo.23433822; program-level protocol record: 10.5281/zenodo.23433826). Pre-registrations, ledgers, analysis scripts and the nine-item checklist template support full reproduction of every claim in the paper.}

\ethics{No human participants or personally identifying data were involved: the underlying studies are simulations over synthetic and public, de-identified data. The protocol-repair taxonomy concerns the auditing of changes to a pre-registered plan in software-intensive, verifier-bound research; we deploy no discovered rule.}

\aucontribute{P.~C. and N.~Z. contributed equally to this work. P.~C.: conceptualization, supervision, project administration, writing -- review and editing. N.~Z.: formal analysis, writing -- review and editing. S.~Z.: conceptualization, methodology, software, investigation, data curation, formal analysis, writing -- original draft. All authors reviewed and approved the final version and agree to be held accountable for the work performed therein.}

\funding{This research received no specific grant from any funding agency in the public, commercial or not-for-profit sectors.}

\competing{The authors declare no competing interests.}

\disclaimer{During the preparation of this work the authors used a large-language-model assistant to improve language and readability; the authors reviewed and edited the content and take full responsibility for the published article.}

\end{document}